# AN IMPROVED U-NET MODEL FOR OFFLINE HANDWRITING SIGNATURE DENOISING

*Wanghui Xiao*

*School of Computer Science and Technology, Chongqing University of Posts and Telecommunications, Chongqing, China.*
**Email: xiaowanghui007@126.com*

## Abstract

Handwriting signatures, as an important means of identity recognition, are widely used in multiple fields such as financial transactions, commercial contracts and personal affairs due to their legal effect and uniqueness. In forensic science appraisals, the analysis of offline handwriting signatures requires the appraiser to provide a certain number of signature samples, which are usually derived from various historical contracts or archival materials. However, the provided handwriting samples are often mixed with a large amount of interfering information, which brings severe challenges to handwriting identification work. This study proposes a signature handwriting denoising model based on the improved U-net structure, aiming to enhance the robustness of the signature recognition system. By introducing discrete wavelet transform and PCA transform, the model's ability to suppress noise has been enhanced. The experimental results show that this modelis significantly superior to the traditional methods in denoising effect, can effectively improve the clarity and readability of the signed images, and provide more reliable technical support for signature analysis and recognition.

## 1 Introduction

The application scope of judicial signature handwriting identification is extremely wide, which can involve multiple fields such as document identification and identity confirmation. Through signature handwriting identification, the authenticity of the document can be accurately judged and the author's identity can be determined, which plays an extremely important role in maintaining social fairness and justice. The formation of handwriting identification opinions mainly relies on the detailed comparison of handwriting samples and handwriting samples. The quality of the samples directly determines the accuracy of the identification opinions. However, in actual situations, handwriting samples usually come from image materials of various historical contracts or archival materials. These images are often mixed with a large amount of interfering information, which brings huge challenges to handwriting identification work. Therefore, removing the noise interference information in the handwriting sample images and obtaining clear and accurate signature handwriting has become an important link in the signature handwriting identification work.

Generally speaking, the acquisition of signature handwriting samples mainly comes from historical archival materials and previous contract signatures. During the process of acquisition and transmission, signature handwriting is bound to introduce different types of noise. Noise can lead to the loss of clarity and details in an image, making it blurry and reducing its contrast, thus making it difficult to distinguish the details in the image. This not only affects subsequent handwriting sample processing such as edge detection and image enhancement, but also influences the results of handwriting identification. Therefore, reducing the noise of handwriting samples and improving the quality of sample images are objective demands in reality, and the significance of handwriting image denoising is of great importance.

Early denoising methods mainly utilized the prior knowledge of images for denoising, among which the BM3D modelwas recognized as one with better effects. The BM3D modelnot only absorbs the denoising method in the wavelet transform domain, but also integrates the idea of similar block matching in NLM, achieving a better denoising effect. In recent years, deep learning has developed rapidly in the field of computer vision, and image denoising algorithms based on deep learning have also been deeply studied. In 2017, Zhang et al. accelerated the training process by using residual learning and batch normalization techniques and proposed a Denoising Convolutional Neural Networks (DnCNN) algorithm, which introduces the convolutional neural network (CNN) into the field of image denoising for the first time and performs denoising under the basic assumption that the residuals are related to the Gaussian error prior. In 2018, the Fast and Flexible Denoising Networks (FFDNet) modeltook an adjustable noise level graph as the input and effectively eliminated variant noise with the help of the specified non-uniform noise level graph. It shows outstanding denoising performance on synthetic and real-world noise images.The Convolutional Blind Denoising Networks (CBDNet) modelfurther realizes the network generalization of the blind denoising network modelfor real images through the noise estimation sub-network of asymmetric learning. Furthermore, Zhang et al. also trained a group of fast and effective Image Restoration Convolutional Neural Networks (IRCNN) with the aid of variable splitting technology, and integrated IRCNN into the model-based optimization method. Thus, a significant image denoising effect was achieved. At the same time, IRCNN was also applied to other low-level visual tasks, providing a fast solution. These denoising methods mainly adopt effective network structures and datasets for training to obtain the reconstructed image signals.In 2023, Shen Nan introduced U-Net into image denoising. As a classic fully convolutional network, U-Net's hierarchical structure can extract the feature information of images at different resolutions to enrich the spatial information and has achieved good denoising effects. However, its single network structure leads to a large amount of redundant computations during the processing, thereby causing the loss of spatial information.

# 2 Related Work

*2.1 U-NET Network*

The U-net network structure was proposed by Ronneberger O et al. in 2015 [39]. Its network structure is shown in Fig.1.

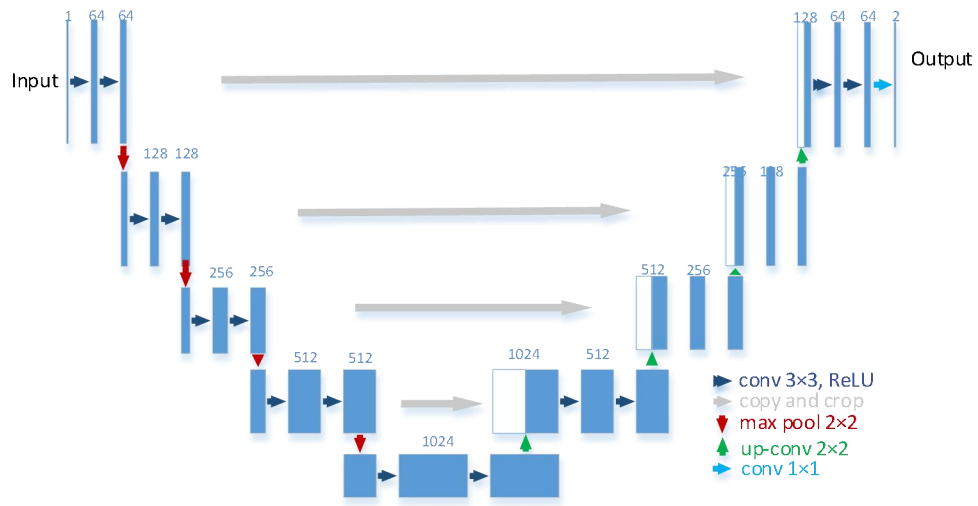

Fig. 1. Original U-Net network structure.

The overall structure of U-NET presents a U shape, consisting of two symmetrical parts: downsampling and upsampling. The downsampling part is composed of continuous convolutional layers and pooling layers, which are used to extract image features and reduce the image size. The upsampling part is composed of consecutive deconvolution layers and convolutional layers, which are used to restore the reduced image to its original size and perform pixel-level classification using the features extracted in the previous stage.

*2.2 Wavelet Transform and Inverse Wavelet transform*

Wavelet transform is an analytical method that decomposes signals or images into components of different scales and frequencies, and can effectively capture the local features and changes of images. Wavelet transform can solve the problem that the window size does not change with frequency. Through scaling and translation operations, the image signal is refined at multiple scales, thereby focusing on any detail of the image signal and having a stronger ability to represent local features. By performing inverse transformation on each frequency band obtained through wavelet transform, the complete information of the image can be synthesized layer by layer.

# 3 Proposed Method

*3.1 Overall Network Architecture*

This study proposes an improved feature fusion denoising method based on U-net network. The overall framework of the network is shown in Fig.2. This network performs multi-scale feature extraction and fusion on the image at the feature extraction layer, thereby enhancing the ability to suppress noise. In the U-net structure, we replace the downsampling with the combination of wavelet transform and Principal Component Analysis, and replace the upsampling with the inverse wavelet transform to enhance the feature extraction ability of the network. Aiming at the problems of inability to fully utilize features and high computational cost in multi-type noise denoising, we combine the U-net structure with various transform analyses and propose a transform analysis image denoising modelbased on the U-net structure, aiming to fully utilize image features to improve image clarity, reduce computational cost and enhance denoising effect.

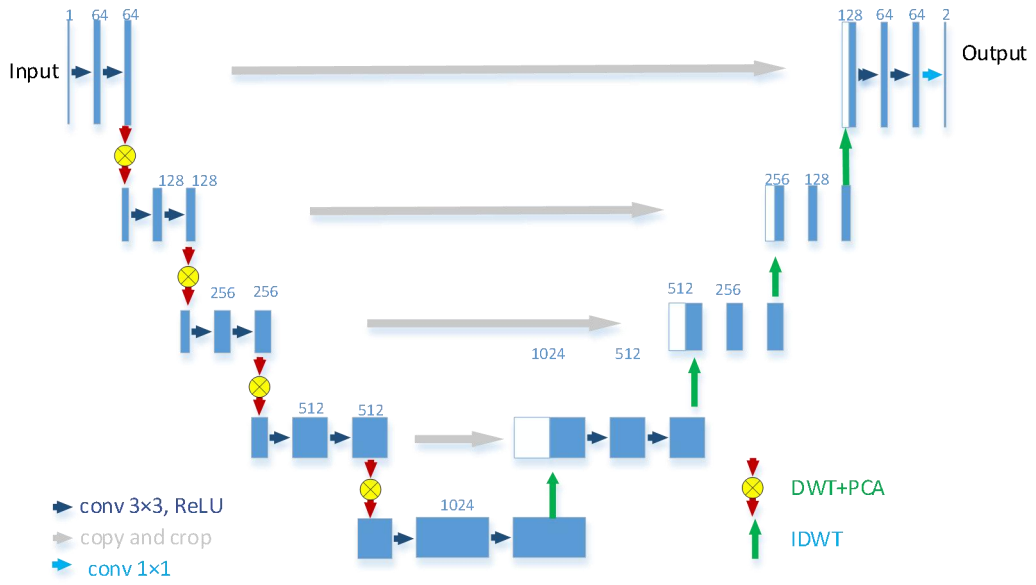

Fig. 2. Improved U-Net network structure.

*3.2 Feature Extraction*

The extraction and utilization of features is one of the important components of neural network denoising. In the proposed network model structure, a noisy image is input, and then feature extraction is carried out through wavelet transform and PCA to obtain relatively obvious shallow features. After that, the above process is repeated to enable the network structure to fully learn the specific features of the noisy image. Finally, in the image reconstruction structure, the inverse wavelet transform method is adopted to reconstruct the image. This is done to achieve the purpose of noise reduction. In the feature extraction structure, the denoising network proposed in this study first performs downsampling processing on the input image. The downsampling methods are discrete wavelet transform and PCA transform based on SVD. The input image is first processed into a three-channel image of the same size. Moreover, in order to retain the information of the original image, padding technology is adopted to increase the number of pixels on each edge to prevent the feature map from being too small and difficult to extract features in the deep network structure.

*3.3* Principal Component Analysis (PCA)

In the field of image denoising, PCA is also often used to extract the main features of the input signal for further denoising the image. The noisy images are processed by PCA to obtain the main features in the noisy images. The specific operation is to compress the extracted high-dimensional features through linear transformation, and then output them after transforming the high-dimensional features into low-dimensional features. The advantage of doing this is that it can reduce the computational cost of the model. Moreover, after linear transformation, it will not destroy the original information structure and reduce the amount of information redundancy. In the field of image denoising, the PCA method utilizes this ability to separate some noise signals while retaining most of the original image features. At the same time of separation, each separated feature is independent and unrelated to each other, so it can also improve the model's resolution ability for noise signals.

*3.4 Feature Fusion*

The features are extracted by using the two methods of wavelet transform and PCA to replace the pooling layer in the convolutional neural network. Before performing the convolution operation on the input image, assuming the input is $x$, the DWT transform and PCA transform are first performed on the input. At this time, DWT($x$) and a PCA($x$) after PCA transformation are obtained. The wavelet transform band has the feature information of most images. The PCA method is adopted to optimize all sub-bands to improve the content of effective information in the pictures. Let the input of the *i-th* layer of the neural network be x, then the output after feature fusion is shown in formula (1)

Output = $\alpha$ DWT($x_i$)+ $\beta$ PCA($x_i$) (1)

Where $\alpha$ and $\beta$ are the transformation coefficients of different methods respectively, and this coefficient is determined by the final test effect.

*3.5 Loss function*

The loss function adopts the mean square deviation function to train the network parameters, and its mathematical formula is:

$$L(\theta)=\frac{1}{N}\sum_{i=1}^{N}\|(y_i-x_i)^2-R(y_i;\theta)\|_F^2 \quad (2)$$

where $\theta$ is the parameter of the u-net network, $R(y_i;\theta)$ is the noise graph obtained through network training. $y_i$ is the image with noise, $x_i$ is the image without noise, and $N$ is the training sample.

# 4 Experiments

*4.1 Dataset and implementation details*

In this paper, the author used the CBSD68 dataset and the real handwriting signature dataset for training. The BSD68 dataset contains 68 color images of different sizes, and the handwriting dataset contains 500 signed handwriting images. The selected Batch Size is 4 and the number of training rounds is 200. The strategy of Warmup for the Learning Rate was used in the training. The initial learning rate was 1e-4, the minimum learning rate was 1e-6. The parameters of the network are optimized using the Adam optimizer provided by Pytorch.

*4.2 Evaluation Indicators*

To quantitatively evaluate the performance of the model, this study uses two of the most commonly used image denoising evaluation indicators, Peak Signal-to-noise Ratio (PSNR) and Structural Similarity Index Measure (SSIM).

$$MSE=\frac{1}{H\times W}\sum_{i=1}^{H}\sum_{j=1}^{W}(X(i,j)-Y(i,j))^2 \quad (3)$$

$$PSNR=10\cdot\log_{10}(\frac{MAX_I^2}{MSE})=20\cdot\log_{10}(\frac{MAX_I}{\sqrt{MSE}}) \quad (4)$$

$$SSIM(x,y)=\frac{(2\mu_x\mu_y+C_1)(2\sigma_{xy}+C_2)}{(\mu_x^2+\mu_y^2+C_1)(\sigma_x^2+\sigma_y^2+C_2)} \quad (5)$$

For two image blocks, their SSIM is always less than 1; 1 indicates complete similarity. Among them, $u_x$ and $u_y$ are the average values of all pixels in the image block, and $\sigma_x$ and $\sigma_y$ are the variances of the pixel values of the image.

*4.3 Performance comparison*

TABLE 1 QUANTITATIVE COMPARISONS WITH OTHER STATE-OF-THE-ARTS METHODS.

| Dataset | α,/β | DnCNN (PSNR/SSIM) | FFDNet (PSNR/SSIM) | Bm3D (PSNR/SSIM) | U-net (PSNR/SSIM) | **Ours (PSNR/SSIM)** |
|---|---|---|---|---|---|---|
| CBSD68 | 1/1 | 32.68/0.8947 | 32.55/0.8702 | 31.74/0.8807 | 31.64/0.8902 | **33.13/0.9005** |
| | 0.7/0.3 | 31.23/0.8379 | 30.19/0.8282 | 30.23/0.8179 | 29.19/0.8077 | **31.94/0.8501** |
| | 0.3/0.7 | 30.20/0.8194 | 29.29/0.8039 | 29.20/0.8094 | 29.29/0.8008 | **30.68/0.8334** |
| handwriting | 1/1 | 28.74/0.8557 | 28.46/0.8562 | 27.77/0.8371 | 27.88/0.8439 | **29.44/0.8687** |

| signature | 0.7/0.3 | 27.23/0.8011 | 26.16/0.8088 | 26.04/0.7982 | 26.35/0.8006 | **27.89/0.8333** |
|---|---|---|---|---|---|---|
| | 0.3/0.7 | 25.20/0.8036 | 25.93/0.7849 | 25.18/0.7294 | 26.12/0.7629 | **26.82/0.8114** |

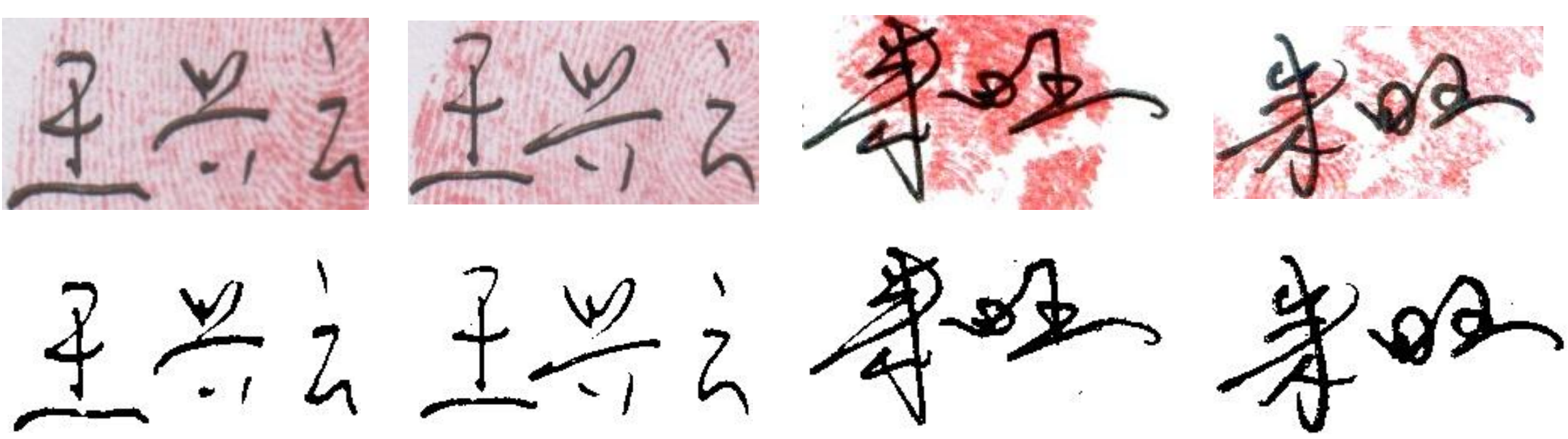

Fig.3 Comparison chart of denoising effects

As shown in Fig.3 and Table 1,compared with previous methods, the prediction accuracy of the method proposed in this paper has been significantly improved. The experimental results show that the model proposed in this study performs better than all other methods in some evaluation indicators of different noise levels in the two datasets, verifying the validity of this model. Compared with other learning-based algorithms (DnCNN, FFDNet ,and Bm3D) , the PSNR of our model has increased by an average of 1.2965dB and the SSIM has increased by an average of 0.0269. Compared with the model based on U-net, the PSNR increased by an average of 1.5717dB and the SSIM increased by an average of 0.0319. The model proposed in this study better retains the edge and texture information of the image, removes noises such as seals on the handwriting, and achieves better subjective visual effects and objective numerical results.

## 5 Conclusion

Aiming at the situation that offline handwriting signature samples contain a large amount of seal and fingerprint noise, an improved U-net structure signature handwriting denoising model is proposed. By introducing discrete wavelet transform and PCA transform, the model's ability to suppress noise is enhanced, effectively improving the clarity and readability of the signed image. Although the improved U-net structure performs well in denoising, future research still needs to focus on the adaptability of the model under different signature styles. By combining diverse datasets, deep learning and traditional methods, a more solid foundation is provided for the advancement of signature recognition technology, achieving broader breakthroughs in signature recognition technology.